\documentclass{article}

\usepackage{arxiv}

\usepackage[utf8]{inputenc} 
\usepackage[T1]{fontenc}    
\usepackage{hyperref}       
\usepackage{url}            
\usepackage{booktabs}       
\usepackage{amsfonts}       
\usepackage{nicefrac}       
\usepackage{microtype}      
\usepackage{lipsum}
\usepackage{graphicx}
\usepackage{natbib}
\graphicspath{ {./images/} }
\usepackage{amsmath} 
\usepackage[ruled,vlined]{algorithm2e} 
\usepackage[caption=false]{subfig}

\usepackage[ruled,vlined]{algorithm2e} 
\usepackage{bm} 
\usepackage{verbatim}

\usepackage{multirow} 

\usepackage{makecell} 

\title{DICE: Deep Significance Clustering \\for Outcome-Aware Stratification}

\author{%
  Yufang Huang  
    \\
  Cornell University\\
  \texttt{yfhuang1992new@gmail.com} \\
  \And 
  Kelly M. Axsom \\
   Columbia University Irving Medical Center\\
   \texttt{kma2161@cumc.columbia.edu} \\ 
   \And 
      John Lee \\ 
   Weill Cornell Medicine  \\ 
   \texttt{jrl2002@med.cornell.edu} \\
   \And 
   Lakshminarayanan Subramanian \\ 
   New York University \\ 
   \texttt{lakshmi@nyu.edu}\\
   \And 
   Yiye Zhang \\
   Cornell University\\
   \texttt{	yiz2014@med.cornell.edu}
}

\begin{document}
\maketitle
\begin{abstract}
We present deep significance clustering (DICE), a framework for jointly performing representation learning and clustering for ``outcome-aware'' stratification. DICE is intended to generate cluster membership that may be used to categorize a population by individual risk level for a targeted outcome. Following the representation learning and clustering steps, we embed the objective function in DICE with a constraint which requires a statistically significant association between the outcome and cluster membership of learned representations. DICE further includes a neural architecture search step to maximize both the likelihood of representation learning and outcome classification accuracy with cluster membership as the predictor. To demonstrate its utility in medicine for patient risk-stratification, the performance of DICE was evaluated using two datasets with different outcome ratios extracted from real-world electronic health records. Outcomes are defined as acute kidney injury (30.4\%) among a cohort of COVID-19 patients, and discharge disposition (36.8\%) among a cohort of heart failure patients, respectively. Extensive results demonstrate that DICE has superior performance as measured by the difference in outcome distribution across clusters, Silhouette score, Calinski-Harabasz index, and Davies-Bouldin index for clustering, and Area under the ROC Curve (AUC) for outcome classification compared to several baseline approaches.  
\end{abstract}


\section{Introduction}

Representation learning~\citep{bengio2013representation,baldi1989neural} and clustering~\citep{xu2005survey} are unsupervised algorithms whose results are driven by input features and priors generally. They are often exploratory in nature, but in certain use cases users have \emph{a priori} expectations for the outputs from representation learning and clustering. In the latter case, having targeted self-supervision in the learning process so as to meet the expectation of the users brings practical value for representation learning and clustering algorithms. This paper proposes deep significance clustering (DICE), an algorithm for self-supervised, interpretable representation learning and clustering targeting features that best stratify a population concerning specific outcomes of interest. Here outcome is a specific result or effect that can be measured. 

DICE is motivated by practical needs in healthcare to develop treatment protocols for subgroups of \emph{similar} patients with \emph{differing} risk levels. The complexity and often the lack of clear clinical practice guidelines warrant the discovery of underlying strata in the data to assist with clinical decision making. An motivating example is creating a safe triage protocol for patients in the emergency departments (ED), where patients present with a wide array of conditions. Two groups of patients may have similar likelihood of a safe outcome but presenting with differing clinical profiles thus needing different interventions in the protocol. Another example, heart failure (HF) is a syndrome that impacts nearly 6 million Americans and is associated with a 50\% 5-year mortality~\citep{ziaeian2016epidemiology}. More than 80\% of individuals suffer from three or more comorbidities~\citep{van2014co}. 
The complexity due to frequent comorbidity or the lack of clear guidelines warrant the discovery of patient subtypes to assist with clinical decision making. For machine learning to assist in this context, it is insufficient to use a classification model to simply classify each patient's outcome. At the same time, using clustering algorithms to identify strata does not guarantee that the stratification is meaningful with respect to the outcome of interest. Existing representation learning, clustering, and classification algorithms serve to cluster patients or classify patients, but few is optimized to jointly achieve these goals. 

\begin{figure*}[htp]
\centering
\includegraphics[width=16cm, angle=0]{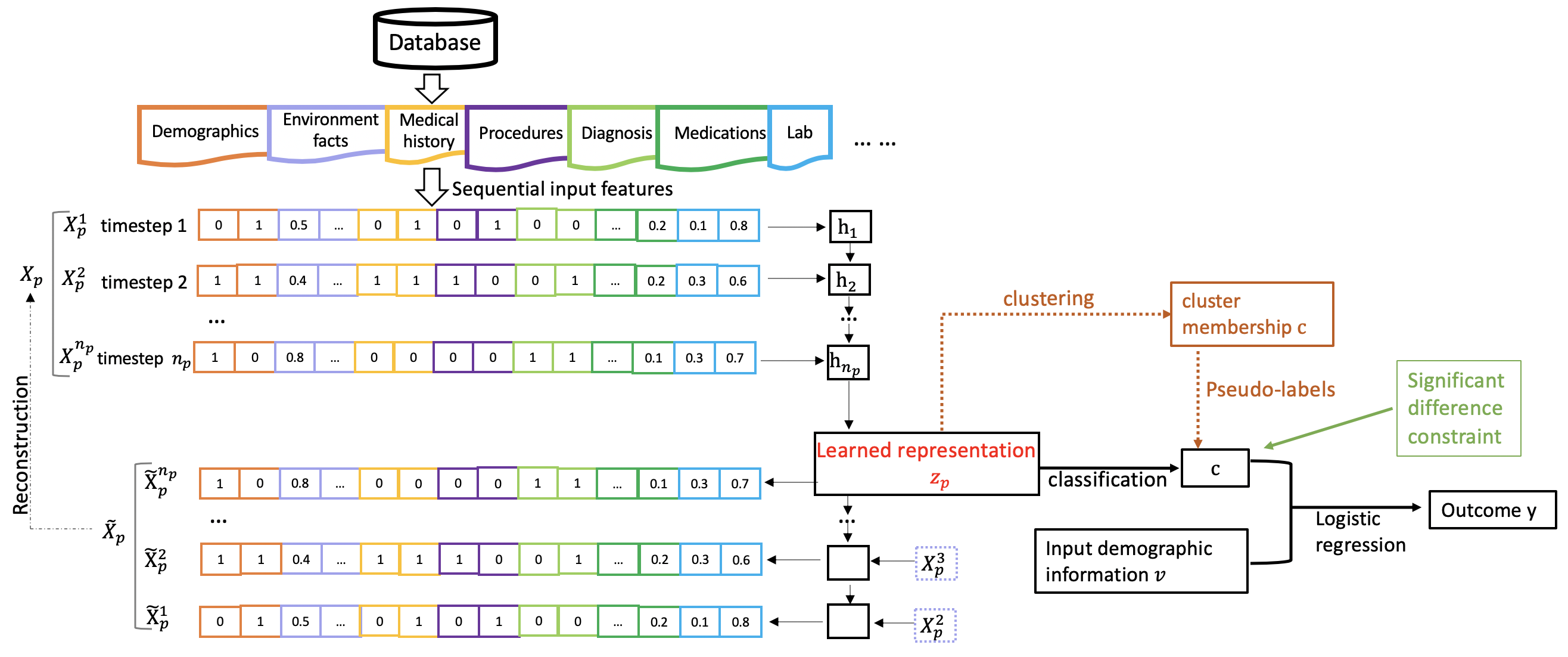}
\caption{The framework of the proposed deep significance clustering (DICE). Clustering is applied to the representation $\mathbf{z}_p$. A statistical significance constraint is explicitly added to ensure the association of the clustering membership $\mathbf{c}$ and outcome $y$, which facilitates the learning of discriminative representations $\mathbf{z}_p$.
} \label{neurips2020_fig:algorithm2_png}
\end{figure*}

DICE, a framework to learn a deep representation and cluster memberships from heterogeneous data was developed in an effort to bridge representation learning, clustering, and classification. Its architecture is illustrated in Fig.~\ref{neurips2020_fig:algorithm2_png}. Representation learning allows us to discover a concise representation from the heterogeneous and sparse health data, which we use to discover latent clusters within a patient population using clustering algorithms. As a way to provide more interpretability of the representation learning and clustering, DICE uses a combined objective function and a constraint that requires statistically different outcome distribution across clusters. The statistical significance is determined using models that are well-understood by clinicians such as regression while adjusting for patient demographics. The combined objective function and constraint serve to force DICE to learn representations that lead to clusters discriminative to the outcome of interest. Furthermore, a neural architecture search (NAS) is designed with an alternative grid search over the number of clusters and hyper-parameters in the representation learning network. The finalized representation and cluster memberships, which represent significantly different outcome levels, are then used as the class labels for a multi-class classification. This is intended to allow new patients to be categorized according to risk-level specific subgroups learned from historic data.

An important distinction between DICE and purely unsupervised, or supervised, algorithms is that DICE learns outcome-aware clusters in an unlabeled population where the outcome-aware clusters are later used to assign risk-levels for future unseen cohort. Previous studies~\citep{zhang2019data} that incorporated statistical significance analyzed it separately after the representation learning process. 
Our paper considers the statistical significance while performing deep clustering as a constraint in an elaborately designed unified framework. To summarize, our approach makes the following key contributions:
\begin{itemize}
    \item We propose a unified objective function to achieve the joint optimization for outcome-driven representation and clustering membership from heterogeneous health data.
    \item We propose an explicit constraint that forces statistical significance of the association between the cluster membership and the outcome to drive the learning.
    \item We utilize a neural architecture search with an alternative grid search for hyper-parameters in the deep significant clustering network.
\end{itemize}

We evaluated DICE on two real-world datasets collected from electronic health records (EHR) data at an academic medical center. Extensive experiments and analyses demonstrate that the DICE obtains better performance than several baseline approaches in outcome discrimination, Area under ROC Curve (AUC) for prediction, and clustering performance metrics including Silhouette score, Calinski-Harabasz index and Davies-Bouldin index.

\section{Related Work}
Clustering is a fundamental topic in the exploratory data mining which can be applied to many fields, including bioinformatics~\citep{lopez2018deep}, marketing~\citep{jagabathula2018model}, computer vision~\citep{yang2019deep} and natural language processing~\citep{blei2003latent}. Due to the inefficiency of similarity measures with high-dimensional big data, traditional clustering approaches, e.g., $k$-means~\citep{macqueen1967some}, finite mixture model~\citep{mclachlan2004finite,wedel1994review} and Gaussian Mixture Models (GMM)~\citep{bishop2006pattern}, generally suffer from high computational complexity on large-scale datasets~\citep{min2018survey}. Also, while mixture models share similar intention as DICE, they further have distribution assumptions on observations~\citep{zhong2003unified}. \citet{jagabathula2020conditional} proposed a conditional gradient approach for nonparametric estimation of mixing distributions. Data transformation approaches which map the raw data into a new feature space have been studied, including principal component analysis (PCA)~\citep{wold1987principal}, kernel methods~\citep{hofmann2008kernel}, model-based clustering~\citep{fraley2002model,zhong2003unified} and spectral methods~\citep{ng2002spectral, von2007tutorial}. However, clustering of high-dimensional heterogeneous data is still challenging for these approaches because of inefficient data representation.

Deep representation learning can be used to transform the data into clustering-friendly representation~\citep{hershey2016deep,xie2016unsupervised,li2018discriminatively,yang2017towards,zhang2019framework}. Parametric t-SNE~\citep{van2009learning} uses deep neural network to parametrize the embedding of t-SNE~\citep{maaten2008visualizing} with the same time complexity of $O(n^2)$, where $n$ is the number of data points. DEC~\citep{xie2016unsupervised} further  relaxes parametric t-SNE with a centroid-based probability distribution which reduces complexity to $O(nK)$ from tree-based t-SNE of $O(n \mathrm{log}(n))$, where $K$ is the number of centroids. Some approaches learn self-supervised representation~\citep{jing2020self,chu2017stacked,caron2018deep}.

Recent deep clustering approaches are learning-based and conduct inference in one-shot, consisting of two stages, i.e., deep representation learning followed by various clustering models. \citet{caron2018deep} jointly learned the parameters of a deep network and the cluster assignments of the resulting representation. DGG~\citep{yang2019deep} further uses gaussian mixture variational autoencoders and graph embedding to improve the clustering and data representation abilities. \citet{yang2017towards} use alternating stochastic optimization to update clustering centroids and representation learning parameters iteratively. Different from~\citet{yang2017towards}, DICE constructs a clustering prediction network and
updates representation learning parameters through self-supervised learning by considering cluster memberships as pseudo-labels of the clustering prediction network. 
Different from \citet{zhang2019framework} adding a constraint on a centroid-based probability distribution, DICE considers statistical significance and proposes a novel constraint added to the cluster membership to obtain statistical significant clustering memberships. 

NAS is a technique to find the network architecture with the highest performance on the validation set. Early NAS conducted architecture optimization and network learning in a nested manner~\citep{baker2016designing,zoph2016neural,zoph2018learning}. These works typically used reinforcement learning 
or evolution algorithms to explore the architecture search space $\mathcal{A}$. 
A recent work decoupled architecture search and weight optimization in a one-shot NAS framework and uses evolutionary architecture search to find candidate architectures after training~\citep{guo2019single}. EfficientNet and EfficientDet~\citep{tan2019efficientnet,tan2019efficientdet} further used grid search to balance network depth, width, and resolution and achieve state-of-the-art results on the ImageNet and COCO datasets respectively~\citep{deng2009imagenet,lin2014microsoft}. We propose an alternative grid search to optimize the number of clusters and other hyper-parameters in the DICE framework.

\section{Method}
Given a dataset $\mathbb{X}=\{\mathbf{X}_1,...,\mathbf{X}_P\}$ with $P$ subjects, we denote each subject as a sequence of events $\mathbf{X}_p=[\mathbf{x}^{1}_{p},\mathbf{x}^{2}_{p},...,\mathbf{x}^{n_p}_{p}]$ of length $n_p$. A multivariate feature vector $\mathbf{x}^{t}_p=[x_{p,1}^t,x_{p,2}^t,...,x_{p,F}^t]\in\mathbb{R}^F$ is the $t$-th instance of subject $p$ in sequence $\mathbf{X}_p$, where $F$ is the number of features at each timestamp. We have an outcome $y_p$ for each subject $p$. Our goal is to stratify $\mathbb{X}$ of $P$ subjects into $K$ clusters while enforcing statistical significance in the association of the cluster membership and the outcome while adjusting for relevant covariates. 

\subsection{Learning representation}
The first step is to transform discrete sequences into latent continuous representations, followed by clustering and outcome classification. The latent representation learning for each subject is performed by an LSTM autoencoder (AE)~\citep{sutskever2014sequence}. The AE consists of two parts, the encoder and the decoder, denoted as $\mathcal{E}$ and $\mathcal{F}$, respectively.  
Given the $p$-th input sequence $\mathbf{X}_p=(\mathbf{x}^{1}_{p}, \mathbf{x}^{2}_{p}, \cdots, \mathbf{x}^{n_p}_{p})$, the encoder can be formulated as 
$\label{encoder}
        \mathbf{z}_p=\mathcal{E}(\mathbf{X}_p;\theta_{\mathcal{E}})
$
, where $\mathbf{z}_p \in \mathbb{R}^d$ is the representation, $d$ is the dimension of representation, and $\mathcal{E}$ is a LSTM network with parameter $\theta_{\mathcal{E}}$ \citep{hochreiter1997long}. We choose the last hidden state $\mathbf{z}_p$ of LSTM to be the representation of the input $\mathbf{X}_p$. The decoder can be formulated as $\tilde{{\mathbf{X}}}_p =\mathcal{F} (\mathbf{z}_p; \theta_{\mathcal{F}})$
, and 
$\mathcal{F}$ is the other LSTM network with parameter $\theta_{\mathcal{F}}$. 
The representation learning is achieved by minimizing the reconstruction error
\begin{equation}\label{AE_optimization}
\begin{aligned}
    &\min_{\theta_{\mathcal{E}},\theta_{\mathcal{F}}}    \mathcal{L}_{AE} = \frac1P\sum_{p=1}^P\|\mathcal{F}(\mathcal{E}(\mathbf{X}_p;\theta_{\mathcal{E}});\theta_{\mathcal{F}})-\mathbf{X}_p\|_{L_2}^2,
\end{aligned}
\end{equation}
where we use $L_2$ norm in the loss.

We employ LSTM networks as encoder and decoder for sequential data, as illustrated in Figure \ref{neurips2020_fig:algorithm2_png}. Our framework can also be used for one-time features (only one timestamp). Multi-layer perceptrons can used as the encoder and decoder for one-time features.

\subsection{Self-supervised learning by clustering}
The obtained representations $\mathbb{Z} = \{\mathbf{z}_p\}_{p=1}^{P}$ can be employed for clustering with $K$ clusters,  
\begin{equation}\label{clustering}
\begin{aligned}
    &\min_{\mathbf{M},\{\mathbf{c}_p\}_{p=1}^P}\mathcal{L}_{clustering}  = \sum_{p=1}^P\|\mathbf{z}_p-\mathbf{M}\mathbf{c}_p\|_2^2 \\
   &\quad \text{s.t.} \quad \mathbf{1}^T\mathbf{c}_p = 1, \; c_{p}^k \in \{0,1\}, \;   \\
 & \quad \quad \forall   \; p\in \{1,2,...,P\},
   \; k\in \{1,2,...,K\},
\end{aligned}
\end{equation}
where $K$ is a hyper-parameter of total number of clusters to tune, $\mathbf{c}_p=[c_p^1,...,c_p^K]$, $c_p^k$ is the cluster membership of cluster $k$, $\mathbf{M}\in\mathbb{R}^{d\times K}$ and the $k$-th columns of $
\mathbf{M}$ is the centroid of the $k$-th cluster. 

To enable fast inference and learn representation with the driven of outcome, we build a cluster classification network for deep clustering based on self-supervision from $\mathbf{c}_p$ in equation~(\ref{clustering}). We employ the clustering results $\{\mathbf{c}_p\}_{p=1}^P$ from \emph{a priori}, such as $k$-means~\citep{macqueen1967some} or Gaussian Mixture Models (GMM)~\citep{bishop2006pattern}, in equation~(\ref{clustering}) as pseudo-labels, and update the parameters of the encoder $\mathcal{E}$ and $\mathcal{F}$. The cluster membership assignment can be formulated as a classification network,  
\begin{equation}
\begin{aligned}
\hat{\mathbf{c}}_p = g(\mathbf{z}_p; \theta_1), \; \quad \min_{\theta_1} \mathcal{L}_1 &= - \sum_{p=1}^P \sum_{k=1}^K c_p^k \mathrm{log} (\hat{c}_p^k),
\end{aligned}
\end{equation}
where $\hat{\mathbf{c}}_p=[\hat{c}_p^1,...,\hat{c}_p^K]$ is the predicted cluster membership from the cluster classification network $g(\cdot ; \theta_1)$, $\theta_1$ is the parameter in the cluster classification network, $\mathcal{L}_1$ is the negative log-likelihood loss for multi-class cluster classification. We will show that deep clustering bridges the representation learning with the following statistical significance constraint related to the outcome.

\subsection{Outcome classification}
After obtaining cluster memberships $\{\mathbf{c}_p\}_{p=1}^P$ for $K$ clusters, we use the cluster memberships and other confounders such as demographics to predict the outcome, formulated as:
\begin{equation}\label{outcome_classification}
\begin{aligned}
&\hat{\mathbf{y}}_p = g([\mathbf{c}_p, \mathbf{v}_p]; \theta_2), \;\\ &\min_{\theta_2} \mathcal{L}_2 = - \sum_{p=1}^P\big( y_p \mathrm{log} (\hat{y}_p) + (1-y_p)\mathrm{log}(1-\hat{y}_p)\big),
\end{aligned}
\end{equation}
where $\mathbf{v}_p$ represents confounders to adjust in testing the significance, $[\cdot, \cdot]$ denotes the concatenation of cluster membership feature and confounders. $g(\cdot ; \theta_2)$ is the logistic regression for the outcome classification, and $\mathcal{L}_2$ is the negative log-likelihood loss for the outcome classification.  

Interpretability is a crucial issue that has not been resolved for the application of deep learning methods in medicine. It's hard to explain why the final outcome prediction is positive or negative for a test case. Using the cluster membership from the learned representation as the input to predict the outcome allows us to infer a broad theme with a set of learned representations, thus
providing more interpretability to the deep representation learning results. Interpretability is further enhanced by enforcing the following statistical significance constraint to the cluster membership $w.r.t.$ the outcome. 


\subsection{Statistical significance constraint}
The main novelty of DICE is the introduction of a statistical significance constraint to the cluster membership $w.r.t.$ the outcome distribution to drive the deep clustering process.
After obtaining cluster memberships $\{\mathbf{c}_p\}_{p=1}^P$ for $K$ clusters, we require that the association between the cluster membership and outcome be statistically significant while adjusting for relevant confounders. 



To quantify the significant difference of cluster $k_1$ and cluster $k_2$ ($k_1\neq k_2$), we use likelihood-ratio test~\citep{david2000applied} to calculate the $p$-value of variable $c^{k_2}$  when considering cluster $c^{k_1}$ as the reference, where $c^k$ refers to the cluster membership belonging to cluster $k$, formulated as, 
\begin{equation}\label{likelihoodratio}
    G_{k_1, k_2} = -2\log \left[\frac{\mathcal{L}_2(g([\mathbf{c}/{\{c^{k_1},c^{k_2}\}},\mathbf{v}];\theta_2), {y})}{\mathcal{L}_2(g([\mathbf{c}/{\{c^{k_1}\}},\mathbf{v}];\theta_2), {y})}\right]
\end{equation}
Then we obtain the $p$-value from Chi-square distribution, denoted as $S_{k_1,k_2}$. Finally, we have a matrix $\mathbf{S}\in \mathbb{R}^{K\times K}$ with $0$ as diagonal elements, and $S_{k1,k2}~(k_1\neq k_2)$ is the $p$-value represent the significance difference of cluster $k_2$ corresponding to reference cluster $k_1$. If all the elements in $\mathbf{S}$ are below a predefined threshold of significance $\alpha$ (equivalently, $G_{k_1,k_2}>\alpha_G$
), we conclude that all the clusters are significantly different with each other related to outcome $y$. In this paper we use $\alpha=0.05$.


In the implementation, we design a mask technique to remove variables of input $\mathbf{c}$, corresponding to cluster $k_1$ and cluster $k_2$, in equation (\ref{likelihoodratio}), then calculate the likelihood ratio $G_{k_1,k_2}$ and add significance constraint to the likelihood-ratio $G_{k_1,k_2}$, that is $G_{k_1,k_2}>\alpha_G, \forall k_1\neq k_2$.

\subsection{Objective function}
We utilize NAS to optimize the network hyper-parameters in the DICE. There are mainly two groups of network hyper-parameters, the hyper-parameter in the clustering and the network hyper-parameters in the representation learning, in the DICE. Basically, NAS conducts two processes iteratively. The first is the neural weights optimization of a given network architecture, which is the network architecture with the fixed number of clusters $K$ and hidden state dimension $d$ in DICE. The second is the neural architecture search process. NAS is conducted in the search phase to select a good combination of hyper-parameters and has no direct association with the cost function of neural weights optimization.

\subsubsection{Optimization of a given network architecture}
We denote our network architecture as $\mathcal{N}(K,d,\theta)$, where $\theta=\{\theta_{\mathcal{E}},\theta_{\mathcal{F}},\mathbf{M},\theta_1,\theta_2\}$ are the weights of network. The neural weights optimization is 
\begin{equation}
\begin{aligned}
\min_{\theta} &\mathcal{L}(\mathcal{N}(K,d,\theta)) \\
=\min_{\theta} &{\lambda}_1 \mathcal{L}_{AE} + \mathcal{L}_{clustering} + \lambda_2 \mathcal{L}_1 + {\lambda}_3 \mathcal{L}_2  + \lambda_4({\alpha}_{G} - G_{k_1, k_2})\\
 \text{s.t.} \quad & \mathbf{1}^T\mathbf{c}_p = 1 \quad  c_{p,j} \in \{0,1\},   \\
   & \forall p\in \{1,...,P\},j\in \{1,...,K\},\\
   & \quad k_1 \neq k_2, \forall k_1, k_2 \in {1, \cdots, K}
\end{aligned}
\end{equation}
where ${\lambda}_1$, ${\lambda}_2$, ${\lambda}_3$, and ${\lambda}_4$ are trade-offs for $\mathcal{L}_{AE}$, $\mathcal{L}_1$, $\mathcal{L}_2$, and the statistical significance constraint. 

We iteratively optimize deep clustering and the other components with the statistical significance constraint. We firstly employ \emph{a priori}, such as $k$-means~\citep{macqueen1967some}, to obtain pseudo-labels for the cluster classification network. Then we can optimize $\mathcal{L}_{AE}$ for the representation learning network, $\mathcal{L}_1$ for cluster classification network, $\mathcal{L}_2$ for outcome classification network, and the statistical significance constraint jointly. The algorithm is elaborated in Algorithm \ref{algorithm}.


\begin{algorithm}[t]\label{algorithm}
\SetAlgoLined
\KwIn{$\mathbb{X}, \{\mathbf{v}\}, K, d$}
\KwOut{$\{\mathbf{z}_p\}_{p=1}^P, \{\mathbf{c}_p\}_{p=1}^P$}
 Initialize the autoencoder of representation learning through $\mathcal{L}_{AE}$\;
Extract representations $\{\mathbf{z}\}$\;
 \For{i = 1 : $n_{iter}$}{
 Optimize $\mathcal{L}_{clustering}$ by $k$-means\; 
 Calculate the cluster membership\;
 Use the cluster memberships as pseudo-labels for cluster classification network in $\mathcal{L}_1$\;
 \For{j = 1 : $n_{epoch}$}{
  Jointly optimize $\mathcal{L}_{AE}$, $\mathcal{L}_1$, $\mathcal{L}_2$, and $G_{k_1, k_2}$ \;
 }
  Extract representations $\{\mathbf{z}\}$\;
 }

 return $\{\mathbf{z}_p\}_{p=1}^P, \{\mathbf{}\mathbf{c}_p\}_{p=1}^P$
 \caption{DICE: Deep significance clustering}
\end{algorithm}

\subsubsection{Architecture search}
We choose the network architecture which is trained on the training set and has the best evaluation performance on validation set, that is 
\begin{equation}
    (K^\star,d^\star) = \operatorname*{argmax}_{K,d} AUC_{val}(\mathcal{N}(K,d,\theta)),
\end{equation}
where $AUC_{val}(\cdot)$ is the AUC score on the validation set. 

\section{Experiments}
We conducted experiments on two datasets and compared against three baseline methods. We also carried out ablation experiments to study the impact of statistical significance constraint of DICE.
\subsection{Experimental setting}
\paragraph{Data}

We used datasets on two patient populations: heart failure (HF) and COVID-19, extracted from electronic health records (EHRs) at an urban academic medical center. The datasets were split into training, validation, and test sets in a $4:1:1$ ratio. 

\begin{itemize}
    \item \textbf{HF}: We included HF patients ($n=1,585$) aged 18 to 89 from years 2014 to 2018 who were treated on the Medicine service. HF was defined by ICD-9/10-CM. 
    The outcome is defined as discharged to home ($36.8\%$). Demographics, medical events (diagnoses, medications and procedures) were included in the data. 
    Events were timestamped by day and concatenated as features. We added normalized days by subtracting initial presentation time into input features. 

    \item \textbf{COVID-19 (AKI)}:
    We included patients aged 18 to 101 who presented to the ED and admitted for COVID-19 disease ($n=1,002$) in 2020.
    COVID-19 was defined by a positive polymerase chain reaction test. The outcome is acute kidney injury (AKI) ($30.4\%$). Age, gender, and laboratory values within 24 hours of ED arrival were included in the data. One-time features for each patient were used. 
\end{itemize}

\paragraph{Baselines}
We compared our method with baseline methods including (1) principal component analysis (PCA) ($k$-means), (2) autoencoder (AE) ($k$-means), and (3) AE w/ classification ($k$-means). For PCA ($k$-means), we merged sequential data into one-time features in HF dataset to learn PCA representations, followed by $k$-means clustering. In AE ($k$-means), $k$-means clustering was applied directly to representations learned from AE~\citep{sutskever2014sequence}. In AE w/ class. ($k$-means), we firstly jointly trained AE and outcome classification with representation learned from AE as the input for outcome classification, then applied $k$-means clustering to the final learned representation. We report the results of these baseline methods of the same hyper-parameters with DICE.  


\paragraph{Training}
We conducted experiments in PyTorch\footnote{https://pytorch.org} on NVIDIA GeForce RTX 2070. We initialized the autoencoder with one epoch training. We set $p$-value $\alpha=0.05$ which leads to $\alpha_G=3.841$, $n_{iter}=60$, $n_{epoch}=1$. The $\lambda_1$, $\lambda_2$, $\lambda_3$, $\lambda_4$ were set as 0.1, 10, 1.0, 1.0, respectively, based on the accuracy on the validation set. It took about 7 minutes to optimize each network architecture. For COVID-19 dataset, the encoder and decoder are set as two-layer fully-connected neural networks with ReLU~\citep{nair2010rectified} activation functions in the intermediate layers.

\begin{figure}[t]
\centering  
\includegraphics[width=12cm, height=5cm]{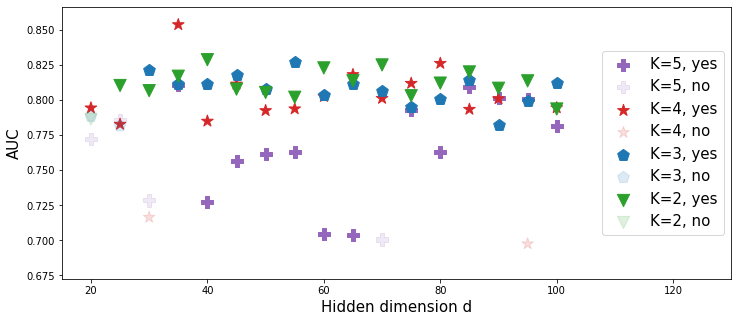}
\caption{The model selection on HF dataset. ``yes'' represents that the architecture network met the significance constraint, and ``no'' otherwise.} \label{neurips2020_fig:tsnek4_withenvir_K4_choosehn}
\end{figure}

\subsection{Results}
We used NAS to choose the best model, then qualitatively compared our method with baselines using clustering and classification metrics. Ablation studies were also conducted to compare performance absent the statistical significance constraint.
\paragraph{Neural network architecture search}

Our search spaces were $\{(K, d) | K\in\{2, 3, 4, 5\}, d \in \{20, 25, ..., 100\}\}$ for the HF dataset and $\{(K,d)|K\in\{2,3,4,5\}, d \in \{10, 11, ..., 20\}\}$ for the COVID-19 dataset, which are set according to the number of features and size of datasets. Figure \ref{neurips2020_fig:tsnek4_withenvir_K4_choosehn} demonstrates the NAS process, with AUC values from the validation set of different neural network architecture on the Y-axis and $d$ on the X-axis. The translucent markers represent that the architectures cannot meet the significance constraint. From Figure \ref{neurips2020_fig:tsnek4_withenvir_K4_choosehn}, we can see that the statistical significance constraint can drive the model towards higher AUC, as also demonstrated in the ablation study described below. 
Maximizing the AUC, the network hyper-parameters $K = 4$, $d = 35$ for the HF dataset, and $K=3$, $d=16$ for the COVID-19 dataset, were chosen as the optimal parameters.

\captionsetup[subfigure]{labelformat=simple, labelsep=colon}
\begin{figure}[htbp]
\centering
\subfloat[DICE.\label{hf:fig:subim1}]{%
  \includegraphics[width=0.24\textwidth]{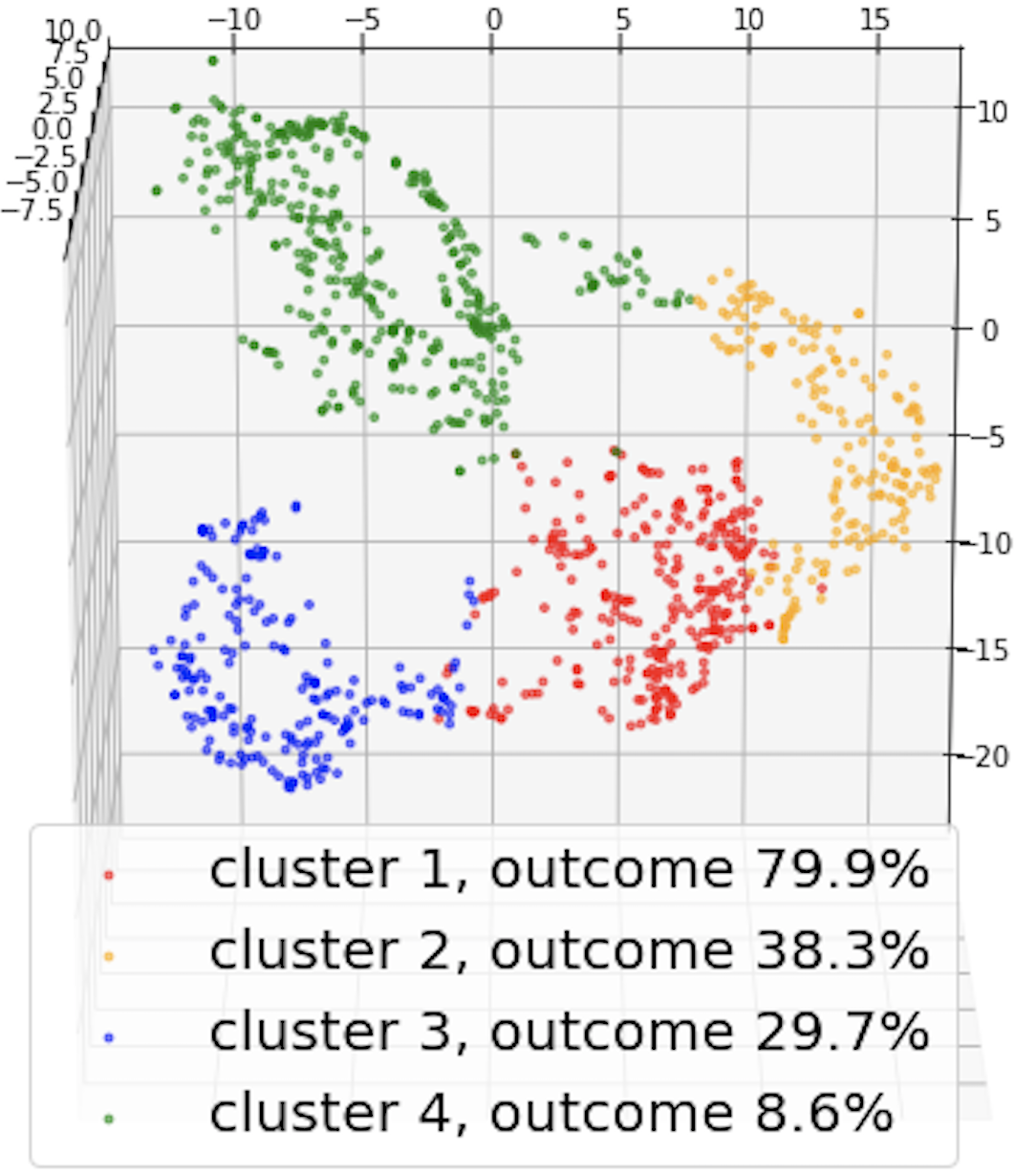}
}
\subfloat[PCA ($k$-means).\label{hf:fig:subim2}]{%
    \includegraphics[width=0.24\textwidth]{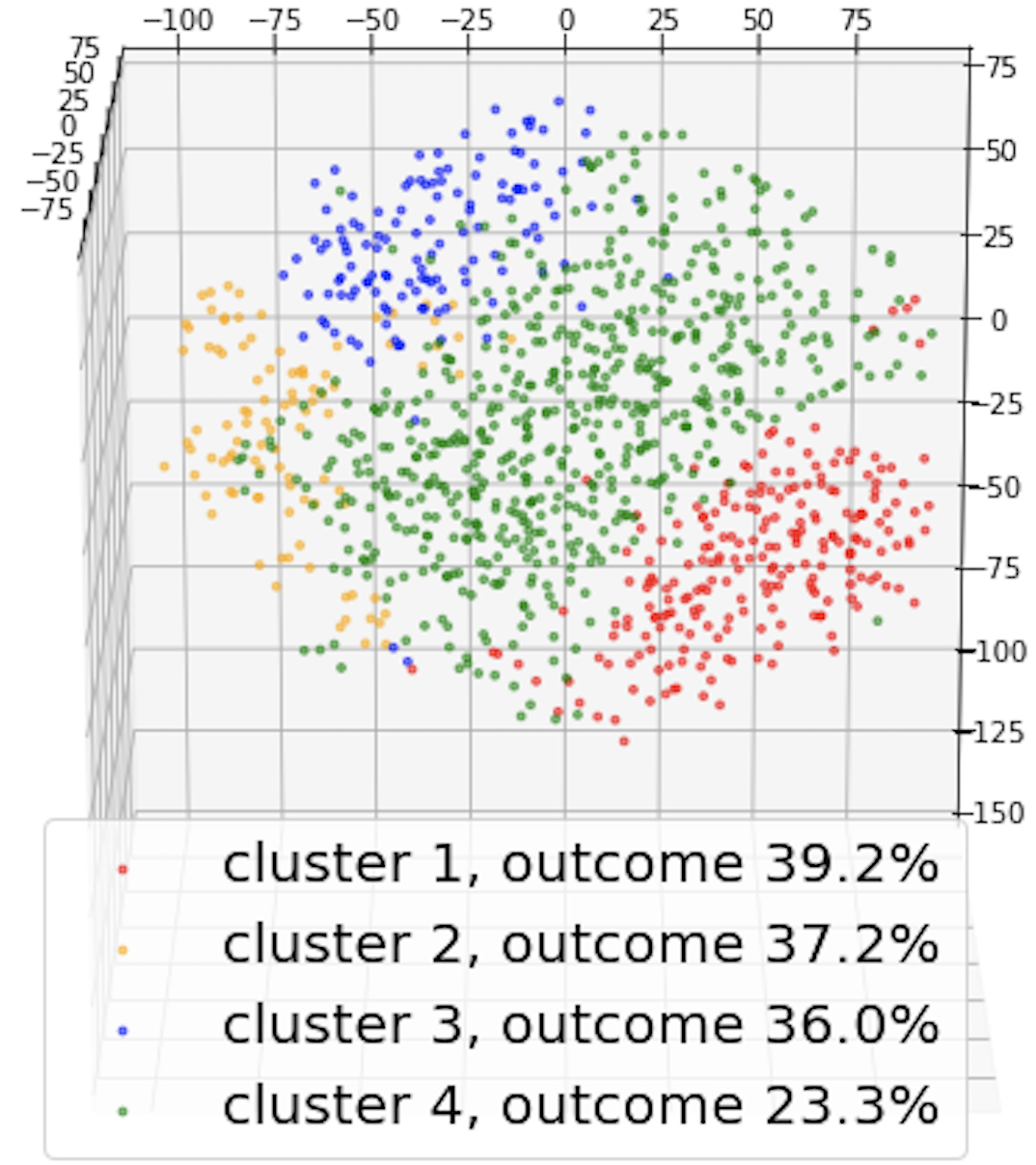}%
}
\subfloat[AE ($k$-means).\label{hf:fig:subim3}]{%
  \includegraphics[width=0.24\textwidth]{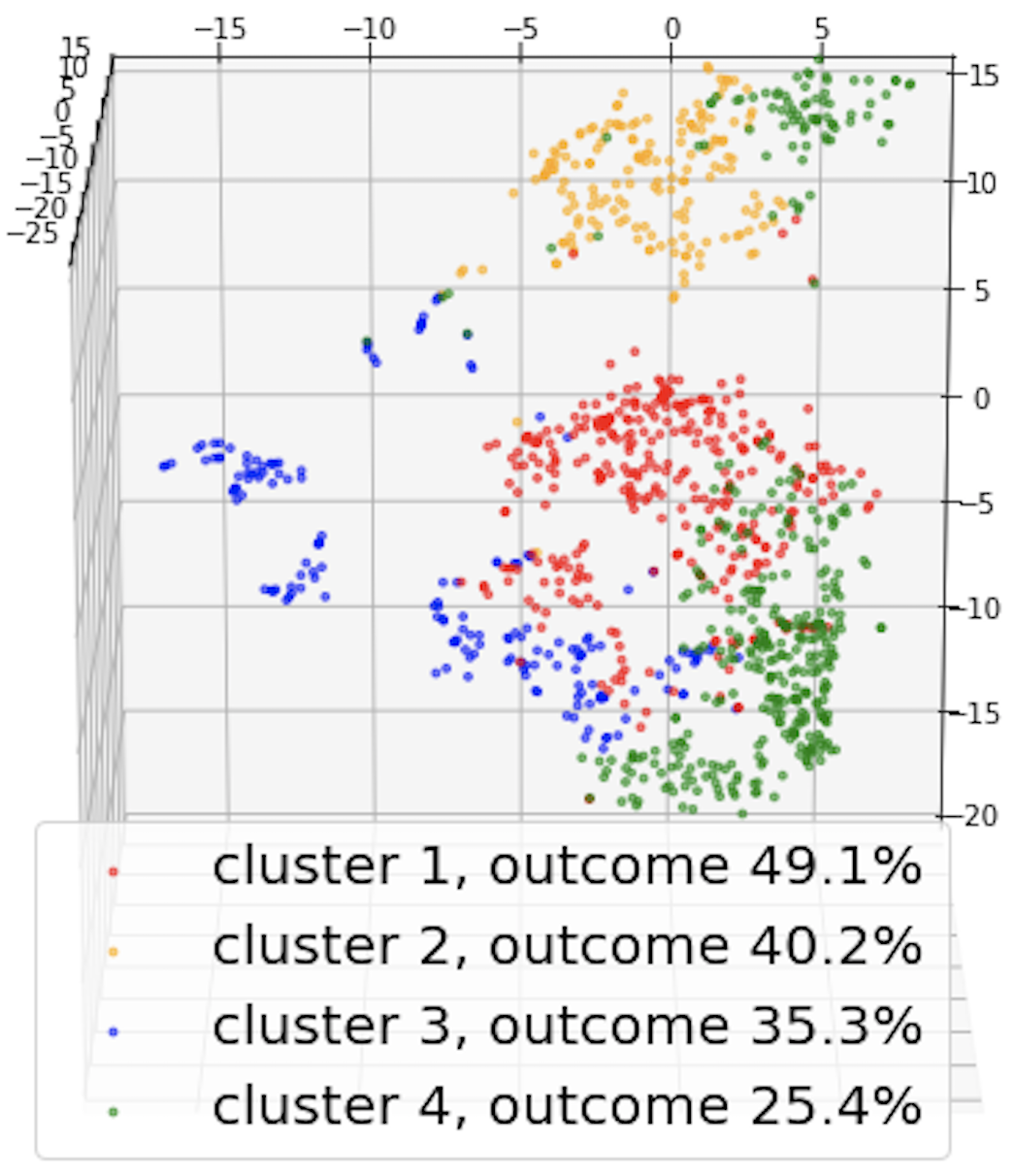}%
}
\subfloat[AE w/ class. ($k$-means).\label{hf:fig:subim4}]{%
    \includegraphics[width=0.24\textwidth]{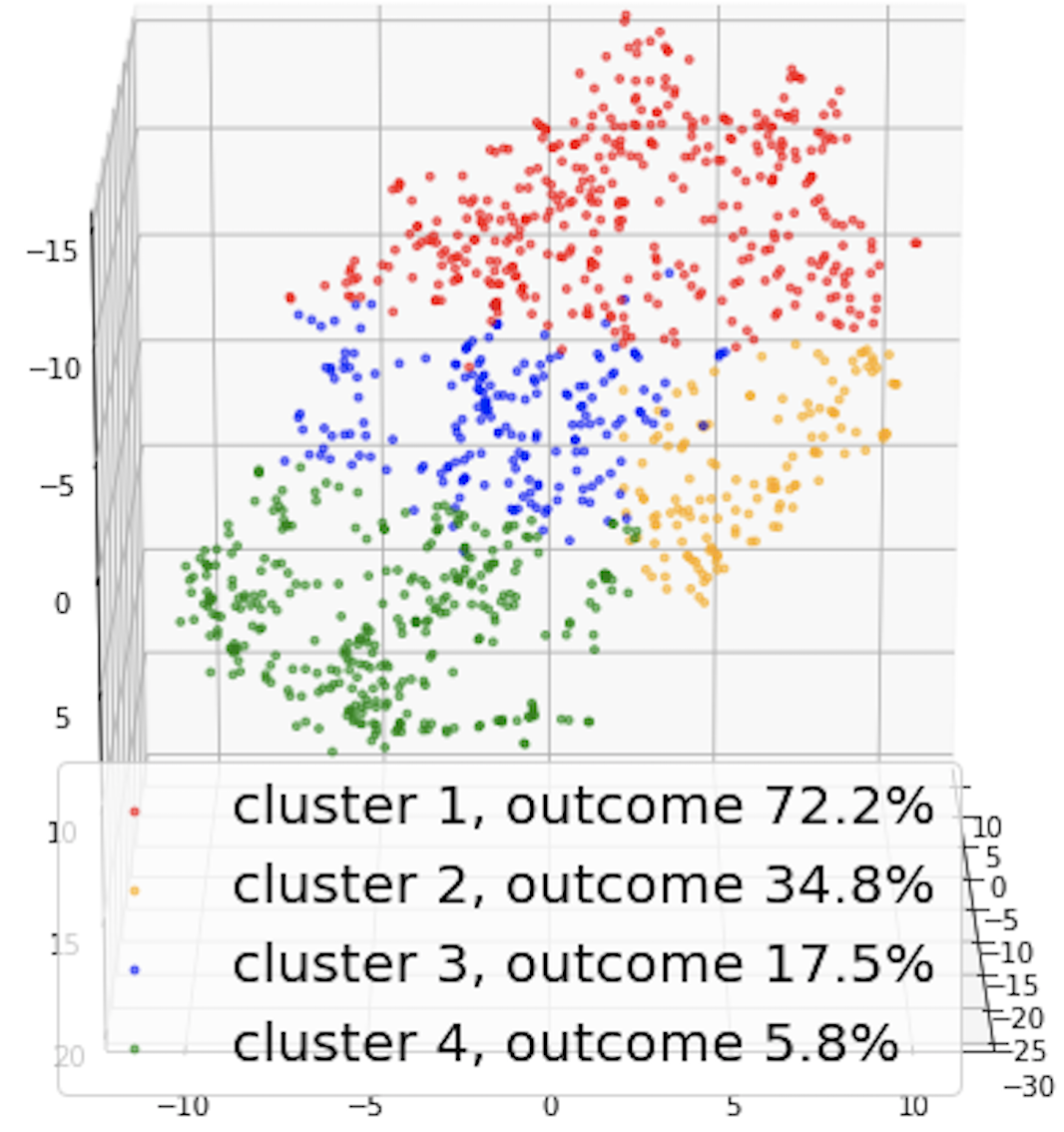}%
}

\caption{Visualization of patient subtyping results by various methods on HF dataset.}
\label{newfigure_tsneD0501_hf}
\end{figure}

\captionsetup[subfigure]{labelformat=simple}
\begin{figure}[htbp]
\centering
\subfloat[DICE.\label{aki:fig:subim1}]{%
  \includegraphics[width=0.24\textwidth]{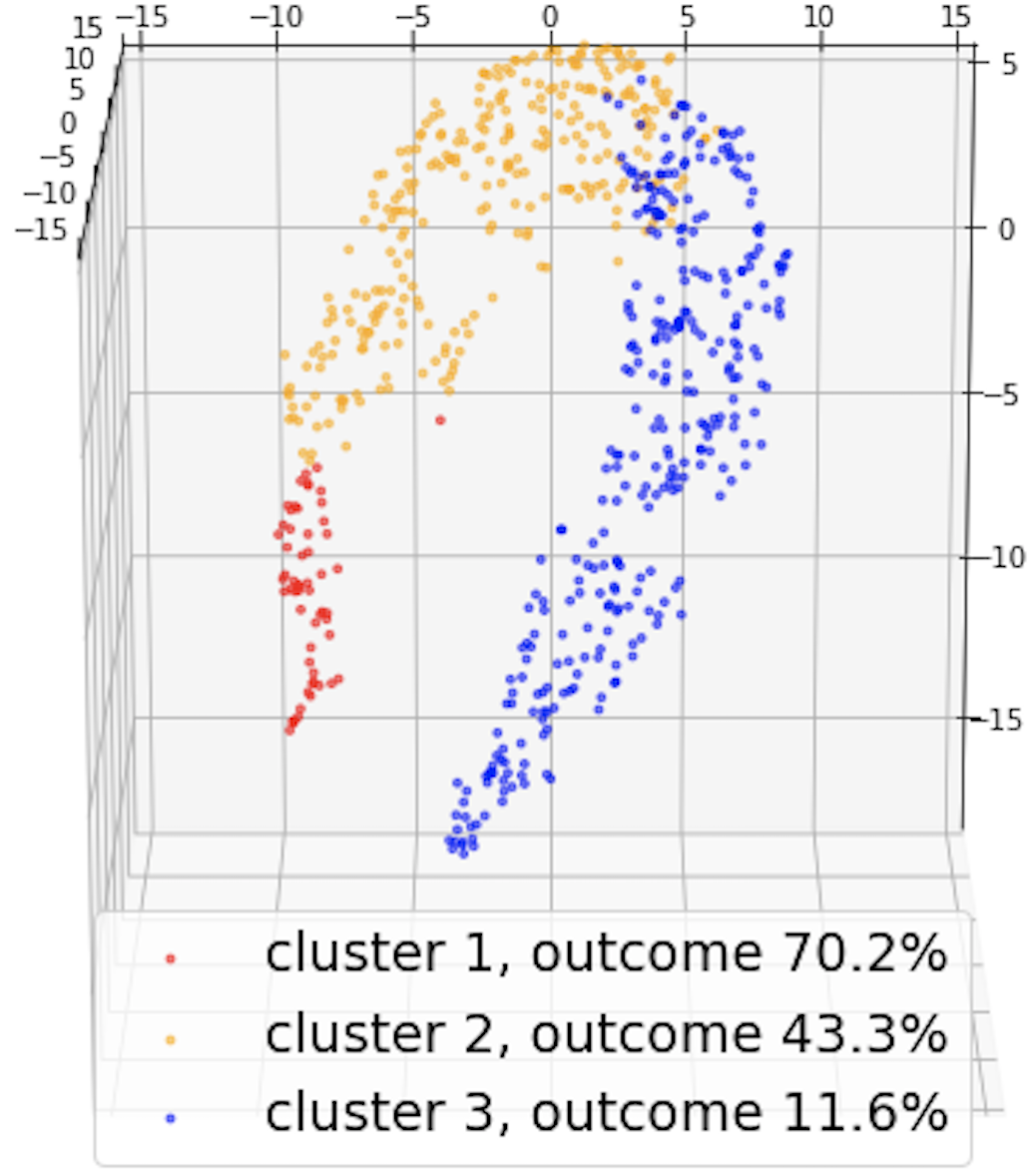}
}
\subfloat[PCA ($k$-means).\label{aki:fig:subim2}]{%
    \includegraphics[width=0.24\textwidth]{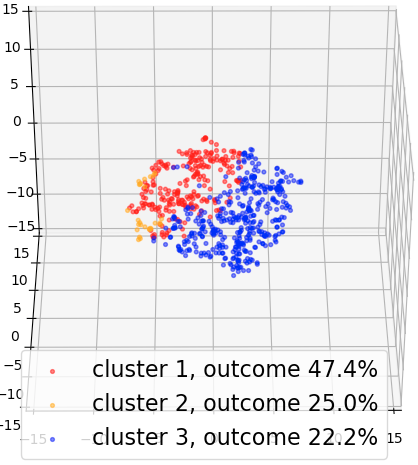}
}
\subfloat[AE ($k$-means).\label{aki:fig:subim3}]{%
  \includegraphics[width=0.24\textwidth]{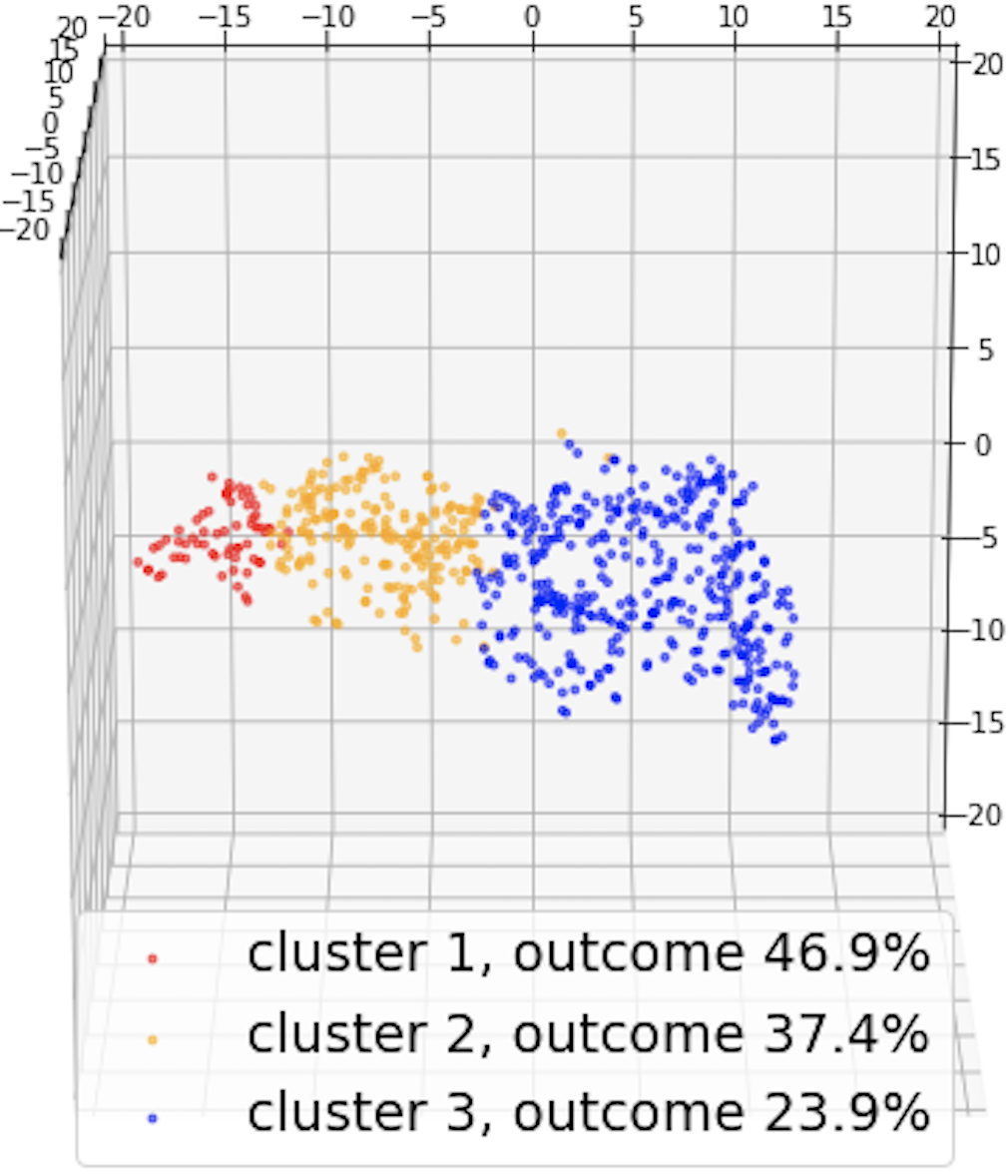}%
}
\subfloat[AE w/ class. ($k$-means).\label{aki:fig:subim4}]{%
    \includegraphics[width=0.24\textwidth]{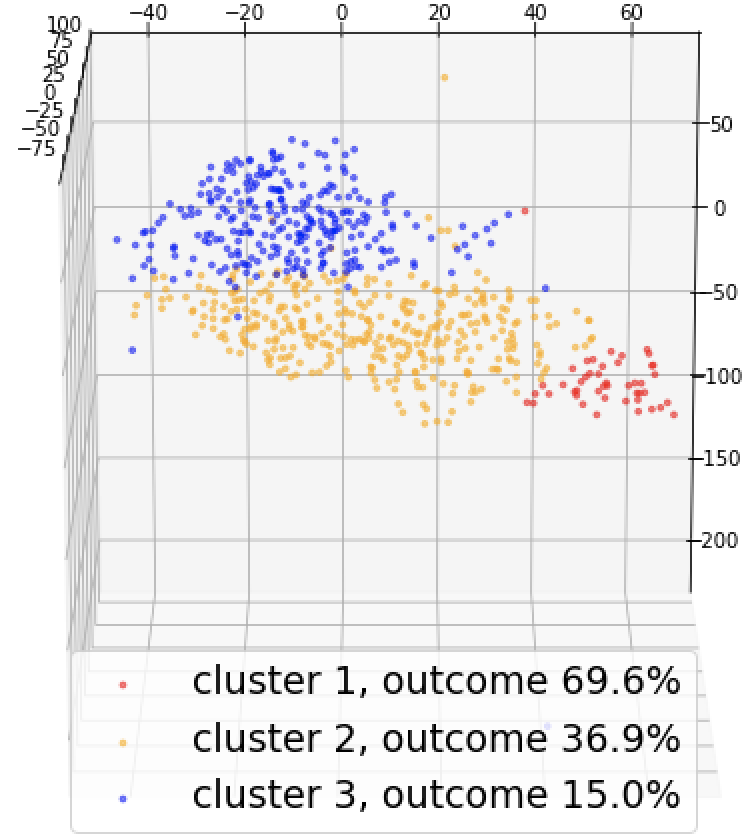}%
}
\caption{Visualization of patient subtyping results by various methods on COVID-19 dataset.}
\label{newfigure_tsneD0526_covid}
\end{figure}

\captionsetup[subfigure]{labelformat=simple}
\begin{figure}[htbp]
\centering
\subfloat[DICE.\label{covid:fig:subim1}]{%
  \includegraphics[width=0.23\textwidth]{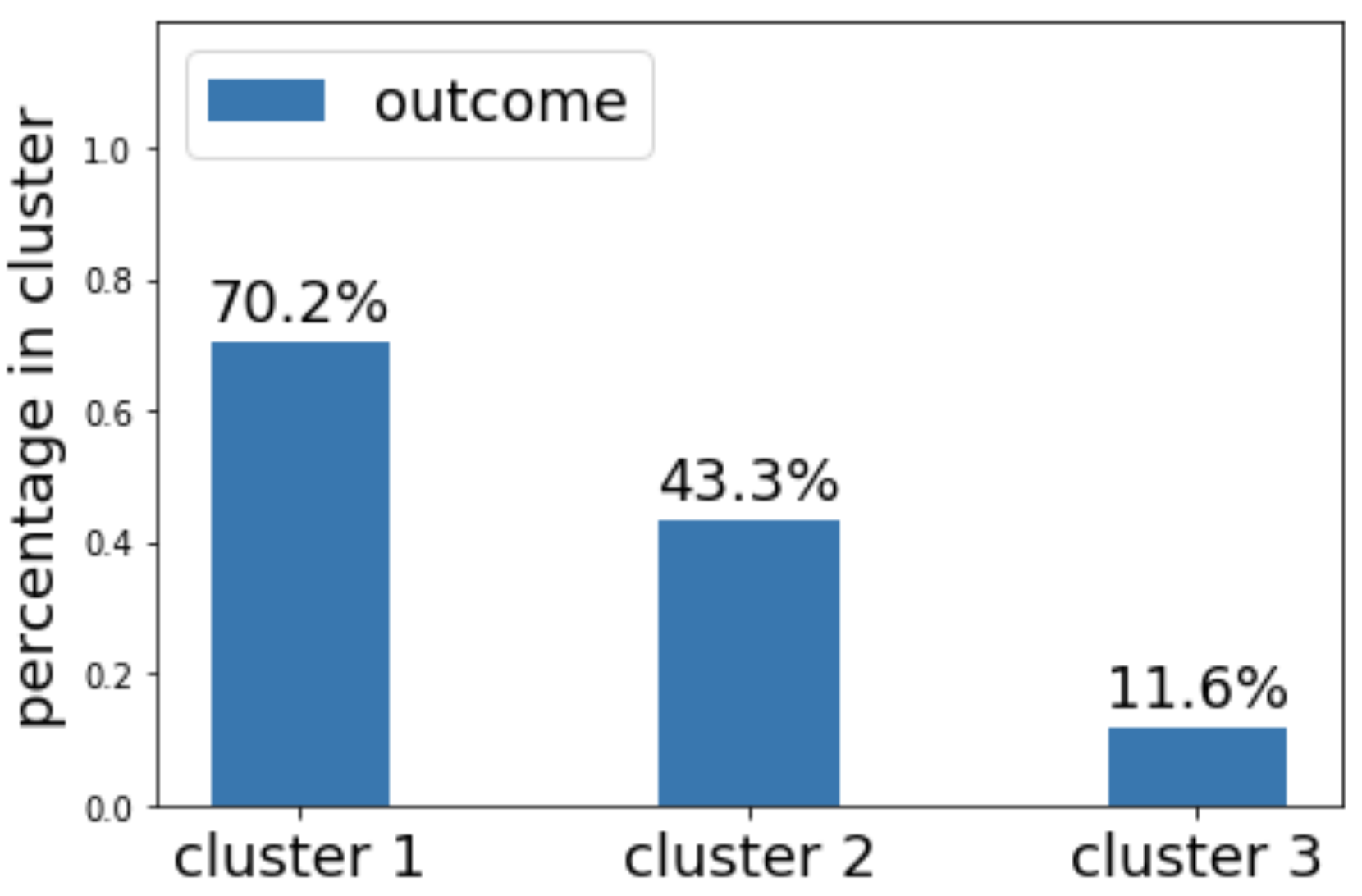}
}
\subfloat[PCA ($k$-means).\label{covid:fig:subim2}]{%
    \includegraphics[width=0.23\textwidth]{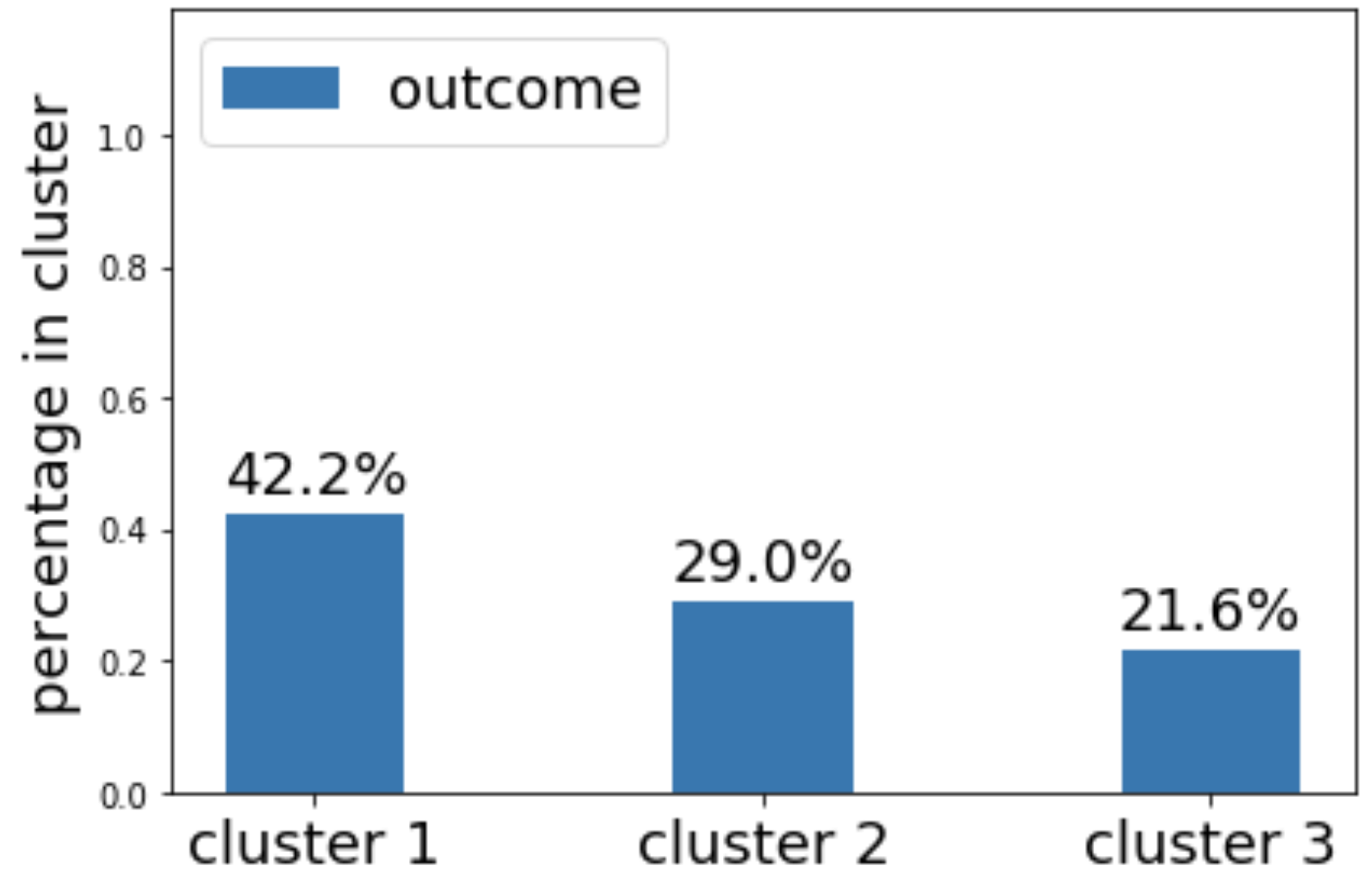}
}
\subfloat[AE ($k$-means).\label{covid:fig:subim3}]{%
  \includegraphics[width=0.23\textwidth]{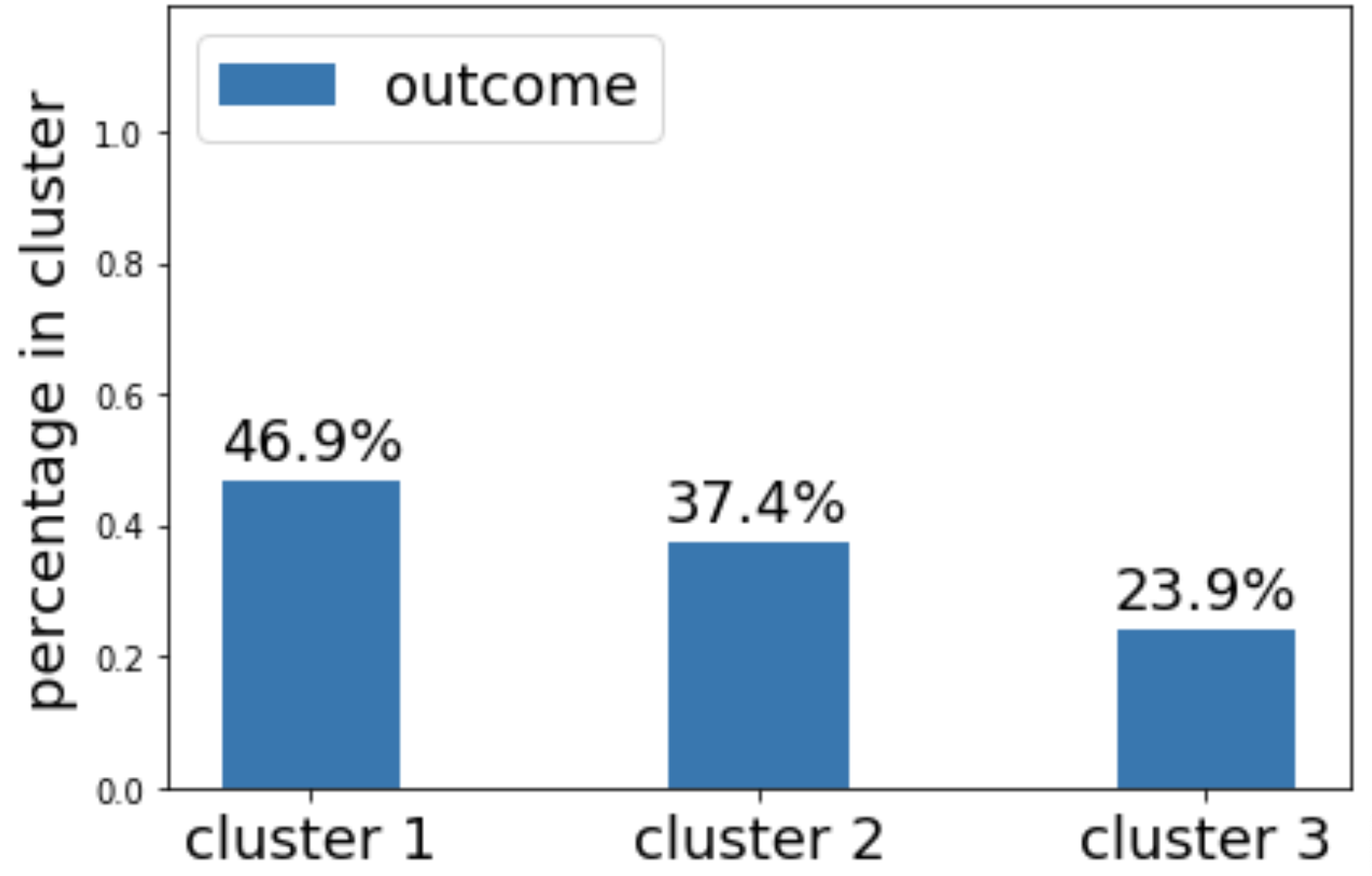}%
}
\subfloat[AE w/ class. ($k$-means).\label{covid:fig:subim4}]{%
    \includegraphics[width=0.23\textwidth]{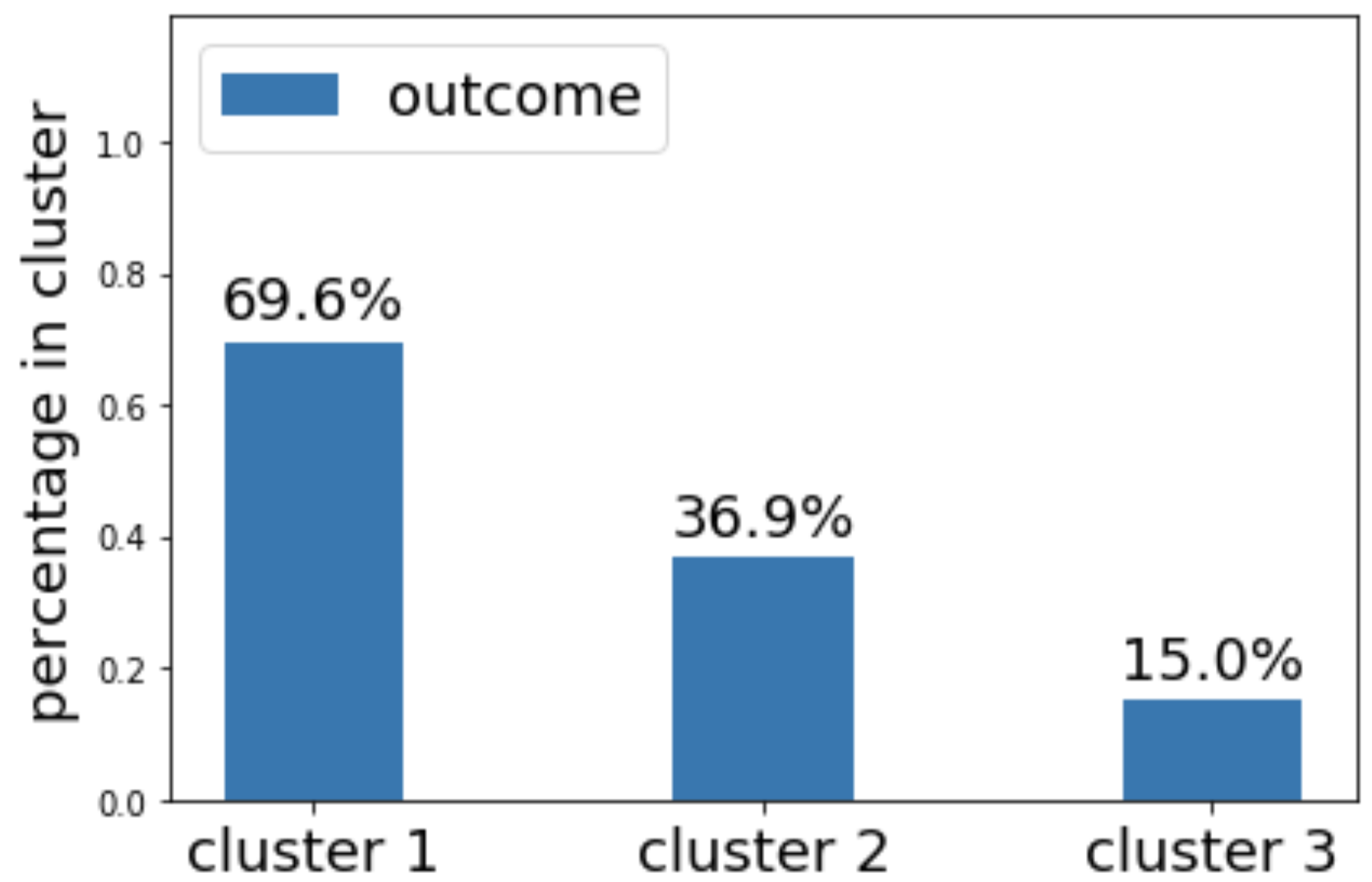}%
}
\caption{Outcome stratification results by various methods on COVID-19 dataset.}
\label{newfigure_tsneD0526_covid_bar}
\end{figure}
\paragraph{Visualization of representation}

For the HF dataset, we demonstrate the clustering results through the visualization of representation in Figure \ref{newfigure_tsneD0501_hf}. Compared with Figure \ref{hf:fig:subim2}, Figure \ref{hf:fig:subim3} and Figure \ref{hf:fig:subim4}, the 4 clusters in Figure \ref{hf:fig:subim1} discovered by DICE displayed tighter separation, with the highest outcome ratio $79.93\%$ in cluster $1$ to the lowest outcome ratio $8.61\%$ in cluster $4$. The baseline AE w/ class. ($k$-means) also discovered 4 clusters with the outcome ratio in each cluster ranging from $72.22\%$ to $5.85\%$, but the clusters are not well separated. PCA ($k$-means) and AE ($k$-means) did not discover clusters with outcomes as clearly separated as DICE, likely because the two baselines are not outcome-driven. Our DICE learns representation through outcome-driven and conducts self-supervised learning with pseudo-labels, 
therefore we can obtain clear outcome risk stratification and well separated clusters at the same time.
Visualizations of patient subtyping results for the COVID-19 dataset are shown in Figure \ref{newfigure_tsneD0526_covid}. DICE again obtained clearer separation between clusters. The outcome stratification results are given in Figure \ref{newfigure_tsneD0526_covid_bar}. From Figure \ref{newfigure_tsneD0526_covid_bar}, we can see that DICE obtained better outcome stratification as measured by the difference in outcome ratio between clusters.

\paragraph{Clustering performance on unseen data}
 The learned cluster membership from historic data can serve as a pseudo-label for unseen data, such that new patients may be classified into one of the risk levels.  The clustering performance on the test set is shown in Table \ref{neurips2020_cluster_evaluation_ontest}. Since the ground truth labels of stratification are unknown, we used Silhouette score~\citep{rousseeuw1987silhouettes}, Calinski-Harabasz index~\citep{calinski1974dendrite}, 
 and Davies-Bouldin index~\citep{davies1979cluster} 
 to evaluate the clustering performance. 
  DICE achieved the best separation across all the three metrics in both HF dataset and COVID-19 dataset.

 \begin{table*}
  \caption{Clustering performance evaluation on the test set. Upper: HF dataset. 
   Lower: COVID-19 dataset.}
  \label{neurips2020_cluster_evaluation_ontest}
  \centering
  \begin{tabular}{cccc} 	\Xhline{2\arrayrulewidth} 
  &\textbf{Silhouette score$\uparrow$}& \textbf{Calinski-Harabasz index $\uparrow$}& \textbf{Davies-Bouldin index $\downarrow$}\\ 	\Xhline{2\arrayrulewidth}  
  PCA ($k$-means)&0.0973&16.0928&2.6093 \\ 
AE ($k$-means)	& 0.2811 & 68.0664&	1.7438\\ 
AE w/ class. ($k$-means)	&	0.3458	&200.0490	& 1.3043	\\ 
DICE	&	\textbf{0.4838}	&	\textbf{212.1706}	&	\textbf{0.8637}		\\ \hline \hline 


 PCA ($k$-means)& 0.1877 & 29.9614 &  1.8403\\ 
AE ($k$-means)	& 0.4622& 162.79197  &0.8413	\\ 
AE w/ class. ($k$-means)	&0.2660	&92.3932	&1.1244\\ 
DICE	&\textbf{0.5141	}	&\textbf{253.5772}	&	\textbf{0.6641}	\\ \hline 

  \end{tabular}
\end{table*}

\paragraph{Outcome classification via learned representation}
We used the learned representation from DICE for outcome classification using logistic regression, as shown in Table \ref{neurips2020_fig:table_compared_with_baseline}. DICE outperformed the baselines in AUC, accuracy (ACC), true positive rate (TPR), false negative rate (FPR), positive predictive value (PPV) and negative predictive value (NPV). The reason DICE had high FPR and low TNR in HF dataset compared to baselines may be explained by the high 
positive case ratio in the HF dataset. 

\begin{table*}
  \caption{Outcome prediction comparison on the test set. Upper: HF dataset. 
  Lower: COVID-19 dataset.}
  \label{neurips2020_fig:table_compared_with_baseline}
  \centering
  \begin{tabular}{c|cccccccc} \Xhline{2\arrayrulewidth} 
	&	\textbf{AUC$\uparrow$} 	&	\textbf{ACC}$\uparrow$ 	&	\textbf{FPR$\downarrow$} 	&	\textbf{TPR$\uparrow$} 	&	\textbf{FNR$\downarrow$} 	&	\textbf{TNR$\uparrow$} 	&	\textbf{PPV$\uparrow$} 	&	\textbf{NPV	$\uparrow$} \\
	\Xhline{2\arrayrulewidth} 
PCA ($k$-means) & 0.773 &0.712 & 0.222 & 0.598& 0.402 & 0.778& 0.611 & 0.769\\
AE ($k$-means)	&	0.712	&	0.697	&	\textbf{0.150}	&	0.433	&	0.567	&	\textbf{0.850}	&	0.627	&	0.721	\\
AE w/ class. ($k$-means)	&	0.818	&	0.765	&	0.251	&	0.794	&	0.206	&	0.746	&	{0.647}	&	0.862	\\ 
DICE	&	\textbf{0.834}	&	\textbf{0.780}	&	0.257	&	\textbf{0.845}	&	\textbf{0.155}	&	0.743	&	\textbf{0.656}	&	\textbf{0.892}	\\ \hline  
\hline 
	

PCA ($k$-means) &0.738 & 0.701& 0.276 & 0.647 & 0.353 & 0.724 & 0.508 & 0.824\\
AE ($k$-means)	 &0.686 & 0.695& 0.285& 0.647& 0.353& 0.716& 0.5& 0.822	\\ 
AE w/ class ($k$-means) &0.734 & 0.689& 0.302& 0.667 & 0.333 & 0.698& 0.493 &0.827\\
DICE	 &\textbf{0.777}&\textbf{0.734}&\textbf{0.263}&\textbf{0.726}&\textbf{0.275}&\textbf{0.737}&\textbf{0.544}&\textbf{0.861}
\\ \hline  

  \end{tabular}
\end{table*}


\paragraph{Fairness on race}
To ensure fairness of the algorithm, we tested DICE within each demographic patient subgroups in the HF dataset. The AUCs for Unknown, Asian, Other, Black, and White are 0.9053, 0.8824, 0.8563, 0.8321, 0.8470, respectively, when \emph{cluster membership} is used as the predictor. The AUCs for Unknown, Asian, Other, Black, and White are 0.8632, 0.8289, 0.7816, 0.8535, 0.8525, respectively, when learned representation is used as the predictor.  

\paragraph{Ablation study}
We conducted an ablation experiment on the HF dataset to gauge the effect of the statistical significance constraint. When we disabled the statistical significance constraint, $2$ clusters, with outcome distributions of $80.1\%$ and $9.01\%$ were chosen by NAS, compared to the 4-level separation in Figure \ref{hf:fig:subim1}. The maximum AUC score with cluster membership as the predictor was $0.8427$ in the ablation study compared to the maximum AUC score $0.8539$. In addition, the percentage of eligible neural network decreased from $82.4\%$ to $64.7\%$ for $K=5$ in the ablation study. These three phenomenons indicate that statistical significance constraint contributes to clearer outcome stratification especially for bigger $K$. 

\section{Conclusion}
We demonstrated DICE using AE for representation learning, followed by a cluster classification network. In the training, we employ $k$-means to generate pseudo-labels to train the cluster classification network, and an alternative grid search in NAS for the optimal network hyper-parameters. In the experiments to discover subgroups of patients in two disease populations: HF and COVID-19, we found that, compared to baseline, DICE better separated the population as measured by clustering indices. The cluster membership from DICE also leads to higher AUC in classifying outcomes, and was further used to assign unseen data into risk-levels.

Future studies will evaluate extension of DICE on multi-class outcomes. In this paper, we conducted experiments on 2 datasets with outcome ratio of roughly 30\%. Future studies will also evaluate DICE on more imbalanced datasets. In addition, the flexibility of the DICE framework will allow alternate methods for representation learning and clustering to be evaluated depending on the needs of the application area. 

DICE joins concepts of deep learning and statistics in medicine to explore clearer presentation of deep learning results. In application, DICE differs from a pure prediction method in that, in addition to predicting individual patients’ risk levels, it simultaneously assigns them into clusters of patients with similar clinical profiles. Thus, DICE also differs from a pure clustering algorithm for its outcome-aware nature in assigning clusters.
Outputs from DICE may be more actionable in alerting healthcare providers of not only high-risk patients, but also providing interpretable insights for subgroup-specific strategies. Beyond HF and COVID-19, DICE may have the potential to be used in other clinical areas to facilitate subtype-specific care and clinical pathways for clinical decision support.

\bibliographystyle{apalike}  

\bibliography{references}
\end{document}